\documentclass{article}

\PassOptionsToPackage{numbers, compress}{natbib}

\usepackage[preprint]{neurips_2019}

\usepackage[utf8]{inputenc} 
\usepackage[T1]{fontenc}    
\usepackage{hyperref}       
\usepackage{url}            
\usepackage{booktabs}       
\usepackage{amsfonts}       
\usepackage{nicefrac}       
\usepackage{microtype}      

\usepackage{amsmath,amsfonts,bm}

\def\eqref#1{equation~\ref{#1}}

\def\1{\bm{1}}

\DeclareMathAlphabet{\mathsfit}{\encodingdefault}{\sfdefault}{m}{sl}
\SetMathAlphabet{\mathsfit}{bold}{\encodingdefault}{\sfdefault}{bx}{n}

\usepackage{hyperref}
\usepackage{url}

\usepackage{wrapfig}
\usepackage{algorithm}
\usepackage{algpseudocode}
\usepackage{multicol}
\usepackage{xcolor}

\hypersetup{colorlinks,citecolor={blue}}

\title{Learning to Generalize to Unseen Tasks\\ with Bilevel Optimization}

\author{%
  Hayeon Lee$^1$,  Donghyun Na$^1$, Hae Beom Lee$^1$, Sung Ju Hwang$^{1,2}$\\
  KAIST$^1$, AITRICS$^2$\\
  South Korea \\
  \texttt{\{hayeon926, donghyun.na, haebeom.lee, sjhwang82\}@kaist.ac.kr}\\
}

\usepackage{amsmath,amsfonts}
\usepackage{amsthm}
\usepackage{amssymb,amsopn}
\usepackage{bm} 

\usepackage{graphicx}  

\def\loss{\mathcal{L}}

\def\S{\mathcal{S}}

\newlength{\widebarargwidth}
\newlength{\widebarargheight}
\newlength{\widebarargdepth}

\newcommand{\eat}[1]{}

\newlength\myindent 
\newcommand{\bx}{\mathbf{x}}

\newcommand{\EE}{{\mathcal{T}}}
\newcommand{\Q}{{\mathcal{Q}}}

\setcitestyle{square}
\setcitestyle{citesep={,}}

\usepackage{amsmath, amssymb, bm, amsthm}
\usepackage{mathrsfs} 
\usepackage{mathtools}
\usepackage{thmtools}
\usepackage{enumitem}
\usepackage{subfigure}
\usepackage{color}
\usepackage{booktabs}
\usepackage{array}
\usepackage{tabularx, booktabs, multirow}
\newcolumntype{M}[1]{>{\centering\arraybackslash}m{#1}}

\usepackage[capitalize,nameinlink]{cleveref}

\usepackage{amsmath}
\makeatletter
\let\oldtagform@\tagform@
\renewcommand{\eqref}[1]{\textup{\oldtagform@{\ref{#1}}}}
\makeatother

\newcommand{\bee}{\begin{eqnarray}}
\newcommand{\eee}{\end{eqnarray}}

\begin{document}

\maketitle


\begin{abstract}
Recent metric-based meta-learning approaches, which learn a metric space that generalizes well over combinatorial number of different classification tasks sampled from a task distribution, have been shown to be effective for few-shot classification tasks of unseen classes. They are often trained with episodic training where they iteratively train a common metric space that reduces distance between the class representatives and instances belonging to each class, over large number of episodes with random classes. However, this training is limited in that while the main target is the generalization to the classification of unseen classes during training, there is no explicit consideration of generalization during meta-training phase. To tackle this issue, we propose a simple yet effective meta-learning framework for metric-based approaches, which we refer to as \emph{learning to generalize (L2G)}, that explicitly constrains the learning on a sampled classification task to reduce the classification error on a randomly sampled \emph{unseen} classification task with a bilevel optimization scheme. This explicit learning aimed toward generalization allows the model to obtain a metric that separates well between unseen classes. We validate our L2G framework on mini-ImageNet and tiered-ImageNet datasets with two base meta-learning few-shot classification models, Prototypical Networks and Relation Networks. The results show that L2G significantly improves the performance of the two methods over episodic training. Further visualization shows that L2G obtains a metric space that clusters and separates unseen classes well.  
\end{abstract}

\section{Introduction}
While deep learning models such as CNNs have been proven effective on multi-class classification~\cite{alexnet,resnet,densenet}, even surpassing human level performances~\cite{imagenet}, such impressive performances are obtained with the availability of large number of training instances per class. However, in more realistic settings where we could have very few training instances for some classes, deep learning models may fail to obtain good accuracies due to overfitting. On the other hands, humans can generalize surprisingly well even with a single example from each class. This problem, known as the few (one)-shot learning problem, has recently attracted wide attention, leading to the proposal of many prior work that aim to prevent the model from overfitting when trained with few instances.

Recently, metric-based meta-learning approaches that learn to generalize over a distribution of task rather than a distribution of a single task~\cite{vinyals2016matching,santoro2016meta,rezende2016one,snell2017prototypical} have obtained impressive performances on the few-shot learning tasks. They tackle the low-data challenge in few-shot learning problems by iteratively training a common metric space over large number of randomly sampled classification problems. Specifically, at each episode, the embedding function that embeds instances onto the metric space, is learned to minimize the distance between the instance embeddings (query) and their correct class embeddings (supports), which are randomly sampled from the entire dataset, based on some distance measure. Matching Networks~\cite{vinyals2016matching}, Prototypical Networks~\cite{snell2017prototypical}, and Relation Networks~\cite{rn} are examples of such metric-based few-shot learning approaches, which are known to perform well and are computationally efficient as well. 
\begin{figure*}[t]
	\centering
	\hfill
	\includegraphics[width=0.8\textwidth]{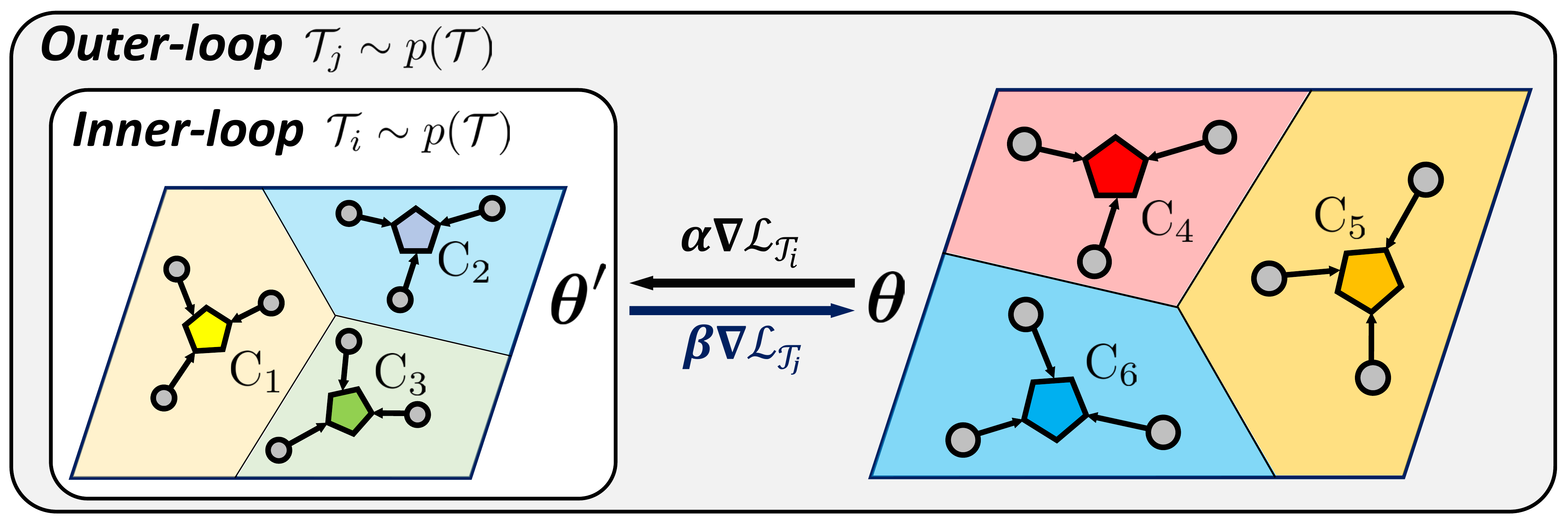}\label{fig:1}
	\hfill
	\caption{\small \textbf{Concept.} Our learning to generalize (L2G) meta-learning framework enforces the learning on the classification task $\mathcal{T}_i$ sampled at each episode, to obtain low classification loss on another classification task $\mathcal{T}_j$.}\label{concept}
\end{figure*}
However, these approaches are limited in that training for each episode optimizes the embedding function to discriminate between classes belonging to a very small subset of the entire class set, without explicit consideration of how this training will affect the classification between classes that are not sampled at this episode. Thus, while the model is trained with the hope that it will generalize to a novel classification task, this effect is only implicitly obtained and cannot clearly separate and cluster the newly given tasks at meta-test time. Using `higher way' or `higher shot' during meta-training to consider classification between more number of classes with larger shots than what they will observe at meta-test time, is one possible solution to improve the generalization performance. Yet, this makes meta-learning expensive since the number of pairs between supports and queries will increase quadratically.

To tackle this limitation of conventional metric-based meta-learning approaches, we propose a novel meta-training framework that explicitly \emph{learns to generalize}, by enforcing the training of the metric space in one task to be also effective for classification between other classes in another task. Specifically, at each training episode, we sample two tasks $\mathcal{T}_i$ and $\mathcal{T}_j$ to solve a bilevel optimization problem~\cite{von2010market}, where we train on task $\mathcal{T}_i$ in the inner loop starting from a common model parameter $\theta$ which not only minimizes the loss on this task but should also minimize the loss on an unseen task $\mathcal{T}_j$. Please see Figure~\ref{concept} for the high-level concept. While in general it may not make sense to train a task-specific model parameter to generalize to another task; however in case of metric-learning approaches, this is possible since the model learns a generic space for classification. With this regularization enforcing the learning on each task to generalize to another, the model learns generic knowledge useful for classification. For example, this will learn that instances for each class should be well clustered in the metric space, in order to classify well between instances from unseen classes.  Since our model requires the task-shared initial parameters to generalize well, it fits well to metric-based meta-learning framework whose solution obtained during meta-training should generalize to unseen tasks without any further training. 

Our meta-learning to generalize (L2G) framework is simple yet effective, and can generalize to any metric-based meta-learning approaches regardless of the specific model details. We validate our L2G framework on benchmark datasets, namely mini-ImageNet and tiered-ImageNet with Prototypical Networks and Relation Networks as the base model. The results show that the L2G framework significantly improves the performances of the both meta-learning models over episodic training. Further visualizations of the learned metric space show that this improvement comes from our model's ability to obtain well-clustered and separable space even on unseen classes.

In summary, our contribution is twofold:

\begin{itemize}
\item We propose a novel and generic meta-learning framework which we refer to as \emph{learning to generalize (L2G)}, that trains a metric-based meta-learning model to explicitly generalize to a different classification task from a sampled classification task at each episode.  

\item We validate our model on two benchmark datasets with two base metric-based meta-learning models, whose results suggest that our L2G framework can significantly improve the generalization performance, regardless of the base model, by obtaining a well-separable metric space on unseen classes.

\end{itemize}

\section{Related Work}
\paragraph{Meta-learning}
Meta-learning~\cite{Thrun1998} is an approach to generalize over tasks from a task distribution, rather than over samples from a single task. Meta-learning approaches could be categorized into memory-based, metric-based, and gradient-based appraoches. MANN~\cite{mann} is a memory-based method, which learns task-generic knowledge by learning to store the instance and its correct label into the same slot, and retrieve it later to predict the label of unseen instances. Metric-based meta-learning approaches tackle the task generalization problem by learning a common metric space that could be shared across any tasks from the same task distribution. Matching networks~\cite{vinyals2016matching} proposed to train a model over multiple episodes (tasks), where at each episode the training set for each class is divided into support instances that represent the class and query instances that is trained to have large similarity to the support instances. Prototypical networks~\cite{snell2017prototypical} proposed to use Euclidean distance for the same purpose, constraining the instances from the same class to be clustered around a single class prototype, and Relation networks~\cite{yang2018learning} further learn the distance metric with additional nonlinear transformation. Gradient-based approaches, such as Model Agnostic Meta-Learning (MAML)~\cite{maml}  aim to learn an initialization parameter that enables to quickly adapt to new tasks with few gradient update steps, and F that learns good initialization parameter for fast adaptation. While meta-learning approaches generalize well to new tasks, this was an effect rather than an explicit objective. On the other hand, in our learning to generalize framework, we explicitly aim for generalization.
\vspace{-0.1in}
\paragraph{Meta-learning with bilevel optimization}
Bilevel optimization, a special kind of optimization that forms nested strucutre of optimization problems, where we have an upper-level problem and a lower-level problem~\cite{von2010market}. The lower-level problems or the inner loop, which expects to hand over the feasible candiates for the upper-level optimization, often includes simpler optimization problems in a constrained situation. Recently, MAML~\cite{maml} proposed to leverage this hierarchical optimization technique to obtain an optimal initialization parameter for a variety of tasks, which became popular as it allowed meta-learning of any models for fast adaptation and generalization to new tasks. Li et al.~\cite{li2017meta} and Lee et al.~\cite{mtnet} further provides regularization on these update stages by introducing learnable learning rate and parameter mask. However, all these models update the outer parameter with respect to the same task used for the inner loop. On the other hand, with our model, the update on the inner loop should solve another task at the outer loop, and thus it is more similar to the original concept of bilevel optimization, which largely focuses on transferring generic information from the pre-simulated tasks, whereas the others simply try to obtain parameters for the given task.

\section{Approach}
\label{approach}

\subsection{Problem Definition}
We start by introducing the episodic training strategy which is widely used for solving few-shot classification problem. Since few-shot learning problems suffer from low-data challenge, many existing works~\cite{vinyals2016matching,santoro2016meta,rezende2016one,snell2017prototypical} resort to meta-learning, which trains the model to generalize over a task distribution $p(\mathcal{T})$. This is done by training it over large number of tasks, where at each task we train the model on a randomly sampled tasks $\mathcal{T}_t\sim p(\mathcal{T})$. By training the model over randomly sampled few-shot classification tasks(or episodes) $\mathcal{T}_1, \dots, \mathcal{T}_T$, we expect to obtain generic knowledge for classification that can be utilized to solve any few-shot classification problems.

Formally, at meta-training time, for each episode $t$, the task $\mathcal{T}_t$ consists of $C$ classes are randomly selected from the training dataset. Then for each class $c$, we randomly sample the support set $S_c = \{\bx_{c, n}\}_{n=1}^N$ and the query set $Q_c = \{\tilde{\bx}_{c, m}\}_{m=1}^M$, where each of $N$ and $M$ denotes the number of support and query instance. By aggregating the support and the query set for all classes as $\S = \{S_c \}_{c=1}^C $ and $\Q = \{Q_c \}_{c=1}^C $, we can define an task $\EE$ is as a tuple $(\S, \Q)$, which is for $C$-way $N$-shot classification problem. At meta test time, we are given a task $\mathcal{T}_{test} = (\S_{test}, \Q_{test})$ where $\S_{test}$ and $\Q_{test}$ contain examples of classes unseen during meta-training time. 

\subsection{Embedding-based Few-shot Learning Apporaches}
We briefly describe a generic framework for metric-based few-shot meta-learning methods~\cite{snell2017prototypical, yang2018learning, vinyals2016matching}. The goal of metric learning approaches is to obtain the optimal embedding function $f_\theta$ with parameters $\theta$ for the given series of task $\EE_1,\dots,\EE_T$. We can handle diverse forms of learning objective depending on modeling assumption. For now we simply denote it as a $\loss$. For a given task $\EE_t$, the metric learning apporaches encode each support instance $\bx \in S_c$ and query instance $\tilde{\bx} \in Q_c$ with the embedding function $f_\theta$ for all $c$. Then, the prototype $\rho_c$ for class $c$ is constructed by adding or averaging embedding vectors to represent each class:
\begin{align}
\rho_c = g(\{f_\theta(\bx_{c,1}),\dots,f_\theta(\bx_{c,N})\})
\end{align}
where $g$ is function which generate a prototype from the embedding vectors of the sample instances.
We can construct loss $\loss$ by computing distance between a query instance $f_\theta(\bx)$ and a set of prototypes $\{\rho_c\}$ for all class within the given task. The distance measure $d(\cdot,\cdot)$ could either be a fixed measure, as with Prototypical Networks~\cite{snell2017prototypical} which leverage the Euclidean distance, or could be learned as with Relation Networks~\cite{yang2018learning} that trains the similarity measure between the two instances using a separate network with additional parameters. Then for a given task $\EE_t$ consisting of the support and query set, we minimize the following loss for all query examples $\tilde{\bx}$:
\begin{align}
\loss_{\EE_t}(\{\rho_c\}, f_\theta(\tilde{\bx})))
\end{align}
The loss should be minimized if $d(\rho_c, f_\theta(\tilde{\bx}_c))$, the distance between $\bx_c$ and its correct class prototype $\rho_c$, is minimized and $d(\rho_i, f_\theta(\tilde{\bx}_c))$ is large, where $i\ne{c}$. 

\subsection{Learning to Generalize to Unseen Classes}
Existing embedding-based approaches learn a metric space over large number of episodes, where given a single episode, a classifier tries to correctly classify query instances based on the class prototypes. However, the main limitation of such an approach is that the embedding function is only trained for classifying given small set of classes at a time, which does not consider generalization of the learned embedding to unseen classes. This explains the reason the metric-based methods such as Prototypical Networks converge fast, but at the same time, trained embedding function may be suboptimal with its myoptic optimization process.

To overcome this shortcoming, we propose a framework which enforces a model to \emph{explicitly} learn transferrable meta-knowledge applicable to any tasks. Specifically, we pair two tasks as a single training unit at each iteration, and train the embedding function by constraining the learning in one task to be helpful to another task during meta-training. The intuition behind this approach is that the network learns more generalizable and transferrable internal features rather than task-specific features. Toward this goal, we adopt a bilevel optimization framework that is similar to MAML~\cite{finn2017model}. The goal of MAML~\cite{finn2017model} is to learn most amenable initialization parameters and to reach specific parameters adapted to the given task through gradient-based parameter updates with this bilevel optimization framework. However, our goal is to learn generalizable parameters for unseen tasks by regularizing the first task which considers the task loss on the second task when it performs optimization. Therefore, differently from MAML~\cite{finn2017model}, in the meta-testing time, our model solves the few-shot classification problems with generalized initial parameters, which do not need fine-tuning with gradient updates for the given tasks. Specifically, we learn for the first task $\EE_i$ in the inner loop while constraining it to obtain low loss on the second task $\EE_j$ in the outer optimization loop, such that we update the shared network parameters for the second task with the gradients generated from the loss of the first task. Since learning on the task $\EE_i$ is regularized to work well for the task $\EE_j$, this will allow task $\EE_i$ to learn a transferrable generic knowledge useful for few-shot classification of any given set of classes.

Note that this is only possible with the models where the same model parameter could be used to solve two different tasks. MAML is not compatible with our learning to generalize framework, since the parameters of the softmax classifier for a few-shot classification task will not work for another few-shot classification task with different sets of classes. On the other hand, with metric-based models, a learned metric for one task is readily generalizable to another task without any modification, as it is essentially a task-generic space where we could embed any instances from any classes onto. Thus we use metric-based meta-learning models for our learning to generalize framework. In Table ~\ref{tbl:mini} and \ref{tbl:tiered}, we show that our model outperforms baselines without finetuning since initial parameter is already optimized to minimize classification error on a task consisting of unseen classes. 

Formally, we construct an inner-gradient update step with the model parameters $\theta$:
\begin{align}
\theta' = \theta - \alpha \nabla_{\theta} \loss_{\EE_i}(\{\rho_c\}, f_\theta(\tilde{\bx})) 
\end{align}

In our case, the step size $\alpha$ is a fixed hyperparameter, but we can learn it as done with Meta-SGD~\cite{li2017meta}. We sample the second task $\EE_j$ from a \emph{disjoint} set of classes to the classes in the task $\EE_i$, and compute the loss $\loss_{\EE_j}(\{\rho_c\}, f_{\theta'}(\tilde{\bx}))$ with updated parameters $\theta'$ by encoding all query input and prototypes of $\EE_j$. Then our goal is to meta-learn a $\theta$ that minimizes the following meta-objective:

\begin{align}
\min_{\theta} {\loss_{\EE_j}}(\{\rho_c\}, f_{\theta'}(\tilde{\bx})) = {\loss_{\EE_j}(\{\rho_c\}, f_{\theta-\alpha \nabla_{\theta} \loss_{\EE_i}(\{\rho_c\}, f_\theta(\tilde{\bx}))}(\tilde{\bx}))}
\end{align}

Note that when the model parameters are updated using the gradients computed from the first task $\EE_i$, the model parameters are optimized to perform the classfication of the second task $\EE_j$. This allows the model to effectively learn the generalized classification for unseen classes across episodes. Then we can perform the meta-update as follows:
\begin{align}
\theta = \theta - \beta \nabla_{\theta} {\loss_{\EE_j}}(\{\rho_c\}, f_{\theta'}(\tilde{\bx}))
\end{align}

where $\beta$ is the meta step size. Algorithm 1 describes the detailed steps of our meta-learning algorithm.

\begin{algorithm}[t]
	\caption{Learning to Genearalize to Unseen Tasks}\label{algo:L2G}
	\begin{algorithmic}[1]
		\State $\alpha, \beta$: step size hyperparameters
		\For{each \emph{batch}}
			\For{each \emph{iteration}}
				\State Sample the first task $\EE_i = (\S, \Q)$
				\State Compute the inner gradient loss $\loss_{\EE_i}(\{\rho_c\}, f_\theta(\tilde{\bx})))$ for $\tilde{\bx} \in \Q$ with $\{\rho_c \}$ 
				\State Evaluate $\loss_{\EE_i}(\{\rho_c\}, f_\theta(\tilde{\bx}))$  in Equation (2)
				\State Update the parameters with gradient descent:
				\State $\theta' = \theta - \alpha \nabla_{\theta} \loss_{\EE_i} $
				\State Sample the second task $\EE_j = (\S, \Q)$ for meta-update
			\EndFor
				\State {\color{red} $\theta = \theta - \beta \nabla_{\theta} {\loss_{\EE_j}}$ using ${\loss_{\EE_j}}(\{\rho_c\}, f_{\theta'}(\tilde{\bx}))$ in Equation (4)}
		\EndFor
	\end{algorithmic}
\end{algorithm}

\subsection{Learning to Generalize with Embedding-based Meta-Learners}
While our framework could work with any metric-based meta-learning methods, we apply it to two most popular models, namely Prototypical Networks~\cite{snell2017prototypical} and Relation Networks~\cite{yang2018learning}.

\paragraph{Prototypical Networks~\cite{snell2017prototypical}}
This is a metric-based few-shot meta-learning model which is discriminatively trained to minimize the relative Euclidean distance between each instance and its correct class prototype over its distances to other class embeddings. Each prototype $\rho_c$ is a mean vector of the support instance embeddings $f_\theta(\bx)$:
\vspace{-0.2in}
\begin{align}
\rho_c = \frac{1}{|S_c|}\sum_{\bx \in S_c} f_\theta(\bx)
\end{align}
For distance measure, we use Euclidean distance:
\begin{align}
d(\rho_c, f_\theta(\tilde{\bx})) = ||\rho_c - f_\theta(\tilde{\bx})||^2
\end{align}
Objective function $\loss$ is based on a softmax over distances to the prototypes in the embedding space with the negative log-probability :
\begin{equation}
\begin{aligned}
&\loss_{\theta} = \sum_{c=1}^C \left[ \sum_{\tilde{\bx} \in Q_c} \left[ d\Big(\rho_c, f_\theta(\tilde{\bx})\Big) + \log \sum_{c'=1}^C \exp\Big(-d\Big(\rho_{c'}, f_\theta(\tilde{\bx})\Big)\Big)\right] \right] \label{eq:disc_loss}
\end{aligned}
\end{equation}

\paragraph{Relation Networks~\cite{yang2018learning}}
This model learns a distance metric using a subnetwork on the concatenated vectors of each class prototype and query instance to generate relation scores between them. An embedding function $f_\theta$ of a Relation Network produces features $f_\theta(\tilde{\bx})$ of query instances and sample instances $f_\theta(\bx)$ of each class. Relation Networks create prototypes $\rho_c$ by adding features from sample instances of each class. 
\begin{align}
\rho_c = \sum_{\bx \in S_c} f_\theta(\bx)
\end{align}
After concatenating each prototype and query instance, the relation module with learnable distance measure computes the relation score between each of the $m$ query instances and $c$ class prototypes:
\begin{align}
s_{c,m} = d(concat(\rho_c, f_\theta(\tilde{\bx}_m)))
\end{align}
Relation Net uses Mean Square Error (MSE) for the objective as follows:
\begin{align}
\loss_{\theta} = \sum_{n=1}^{N_t} \sum_{m=1}^{M_t}(s_{n,m} - 1)_{y_n=y_m}^2 + (s_{n,m})^2_{y_n \neq y_m}
\end{align}
where $N_t = C \times N, M_t = C \times M$ and $y$ denotes the labels. The model is trained to match the support and query instances in each class with the relation score that penalizes the incorrect predictions.
\section{Experiment}

\paragraph{Datasets} We validate our framework on two benchmark datasets for few-shot classification.

\textbf{1) \textit{mini}-ImageNet.} This dataset is a subset of a ImageNet~\cite{krizhevsky2012imagenet} which consists of $100$ classes with $600$ examples for each. We follow the dataset split and pre-processing procudure described in ~\cite{ravi2016optimization}, which divides the dataset into 64/16/20 classes for training/validation/test, and resizes original image into $84 \times 84$ pixels.

\textbf{2) \textit{tiered}-ImageNet.} This dataset is also a subset of the ImageNet~\cite{krizhevsky2010convolutional} with $608$ classes. We split the dataset into 351/97/160 classes for train/validation/test set and follow other experiment settings as described in~\cite{ren2018meta}.

\paragraph{Baselines} We compare models trained with our framework against various meta-learning baselines.

\textbf{1) Matching Networks.} This is another metric-based model~\cite{vinyals2016matching} that is trained using episodic training. However, this model leverages cosine distance instead of Euclidean distance.

\textbf{2) Meta-learner LSTM.} An optimization-based meta-learning model which trains an LSTM based optimizer~\cite{ravi2016optimization} over a distribution of tasks during meta-training, which is used to optimize the target problem at meta-test time. 

\textbf{3) MAML} Model-Agnostic Meta-Learning(MAML) model~\cite{finn2017model}, which aims to learn a shared initialization parameters that can adapt to any given tasks with only a few gradient update steps. 

\textbf{4) Prototypical Network} Prototypical Networks~\cite{snell2017prototypical} described in the previous section.

\textbf{5) Relation Network} Relation Networks~\cite{yang2018learning} described in the previous section.

\textbf{6) MAML + X} 
This is a metric-based meta-learning models trained in the original MAML framework, which is trained with bilevel optimization scheme, but is not trained to generalize to another task. In our experiments, we use MAML + Prototypical Networks and MAML + Relation Networks which are basically Prototypical Networks and Relation Networks trained in MAML framework.

\textbf{7) L2G + X} L2G + X denotes our proposed model, which trains the model to generalize well to another task, in a bilevel optimization framework. We implement both L2G + Prototypical Networks and L2G + Relation Networks.

\paragraph{Implementation Details}
We adopt the same network architectures as baselines for fair comparison. The base networks of Prototypical Network~\cite{snell2017prototypical} have four layers of convolutional blocks where each block contains 64 filters, $3 \times 3$ convolution layer, a batch normalization, a ReLU activation function and $2\times 2$ max pooling layer at the end of each block. The base networks of Relation Net~\cite{yang2018learning} have four convolutional blocks as the embedding function and two convolutional blocks and two fully connected layers which are the relation module to compute the relation score. The former two convolutional blocks of Relation Net have the same architecture to the block of Prototypical Network while the max-pooling layers of the latter two blocks are removed.  

We use Adam optimizer~\cite{kingma2014adam} for training in all experiments. We set the update step size $\alpha$ to $10^{-2}$, the number of tasks for the backward propagation to 5 and the initial learning rate to $10^{-3}$ which is multiplied by $0.5$ at every 10K episode for L2G + Proto and 100K episodes for L2G + RN.

\paragraph{$C$-way $N$-shot Classification}

We evaluate our L2G framework first on the conventional $C$-way $N$-shot classification task against relevant baselines~\cite{maml, snell2017prototypical, yang2018learning}. Following ~\cite{snell2017prototypical}, we train and evaluate models over large number of episodes, where at each episode we randomly sample $C$ classes with $N$ supports for each class. For our L2G framework, we generate and sample two tasks at each iteration in meta-training time. Note that the episode used for inner loop does not participate in the outer loop classification, which is clearly different from the "higher way" or "higher shot" setttings described in ~\cite{snell2017prototypical} where they use larger number of sampled classes or examples which participate in the classification. At meta-test time, our model sample only one task for each iteration and classifies between the unseen classes of the given task with the initial model without any further training.

\begin{table}[t]
	\small
	\caption{\small \textbf{\textit{mini}-ImageNet few-shot classification.} The reported numbers are mean and standard errors with $95\%$ confidence interval over $5$ runs. Each run consists of the mean accuracy over $600$ episodes.} 
	
	\begin{center}
		\begin{tabular}{ccccc}
			\hline
			& \multicolumn{2}{c}{5-way} & \multicolumn{2}{c}{10-way} \\\cmidrule{2-5}
			Models & 1-shot & 5-shot & 1-shot & 5-shot \\
			\hline
			\hline
			Matching Nets~\cite{vinyals2016matching} & 43.56 $\pm$ 0.84 & 55.31 $\pm$ 0.73 & - & - \\	
			Meta-learn LSTM~\cite{ravi2016optimization} & 43.44 $\pm$ 0.77 & 60.60 $\pm$ 0.71 & - & - \\	
			MAML~\cite{maml} & 48.70 $\pm$ 1.84 & 63.11 $\pm$ 0.92 & 31.27 $\pm$ 1.15 & 46.92 $\pm$ 1.25\\	
			Prototypical Network~\cite{snell2017prototypical} & 46.14 $\pm$ 0.77 & 65.77 $\pm$ 0.70 & 32.88 $\pm$ 0.47 & 49.29 $\pm$ 0.42 \\
			MAML + Prototypical Network &  46.73 $\pm$ 0.27 &  65.58 $\pm$ 0.32 & 32.31 $\pm$ 0.23 & 47.61 $\pm$ 0.21 \\
			\bf L2G + Prototypical Network &  \bf 50.20 $\pm$ 0.45 &  \bf 66.19 $\pm$ 0.33 &  \bf 33.82 $\pm$ 0.24 & \bf 50.71 $\pm$ 0.12 \\
			\hline
			Relation Network~\cite{yang2018learning} & 51.38 $\pm$ 0.82 & 67.07 $\pm$ 0.69 & 34.86 $\pm$ 0.48 & 47.94 $\pm$ 0.42 \\
			MAML + Relation Network &  52.08 $\pm$ 0.37 &  66.70 $\pm$ 0.28 & 35.07 $\pm$ 0.18 & 51.52 $\pm$ 0.14 \\
			\bf L2G + Relation Network & \bf 52.38 $\pm$ 0.36 & \bf 68.11 $\pm$ 0.15 & \bf 36.11 $\pm$ 0.30 & \bf 51.40 $\pm$ 0.16 \\
			\hline
		\end{tabular}\label{tbl:mini}
	\vspace{-0.15in}
	\end{center}
\end{table}

\begin{table}[t]
	\small
	\caption{\small \textbf{\textit{tiered}-ImageNet few-shot classification.} The reported numbers are mean and standard errors with $95\%$ confidence interval over $5$ runs. Each run consists of the mean accuracy over $600$ episodes.} 
	\begin{center}
	\begin{tabular}{ccccc}
		\hline
		& \multicolumn{2}{c}{5-way} & \multicolumn{2}{c}{10-way} \\\cmidrule{2-5}
		Models & 1-shot & 5-shot & 1-shot & 5-shot \\
		\hline
		\hline
		MAML~\cite{maml} & 51.67 $\pm$ 1.81 & 70.30 $\pm$ 1.75 & 34.44 $\pm$ 1.19 & 53.32 $\pm$ 1.33\\	
		Prototypical Network~\cite{snell2017prototypical} & 48.58 $\pm$ 0.87 & 69.57 $\pm$ 0.75 & 37.35 $\pm$ 0.56 & 57.83 $\pm$ 0.55 \\
		\bf L2G + Prototypical Network &  \bf 53.47 $\pm$ 0.56 &  \bf 72.19 $\pm$ 0.19 &  \bf 37.38 $\pm$ 0.27 & \bf 58.44 $\pm$ 0.23 \\
		\hline
		Relation Network~\cite{yang2018learning} & 54.48 $\pm$ 0.93 & 71.31 $\pm$ 0.78 & 36.32 $\pm$ 0.62 &  58.05 $\pm$ 0.59 \\
		\bf L2G + Relation Network & \bf 56.09 $\pm$ 0.32 &  \bf 71.97 $\pm$ 0.21 & \bf 39.15 $\pm$ 0.34 & 57.82 $\pm$ 0.46 \\
		\hline
	\end{tabular}\label{tbl:tiered}
	\vspace{-0.15in}
\end{center}
\end{table}


\begin{figure*}[t]
	\vspace{-0.1in}
	\centering
	\subfigure[mini-ImageNet]{\includegraphics[height=4cm]{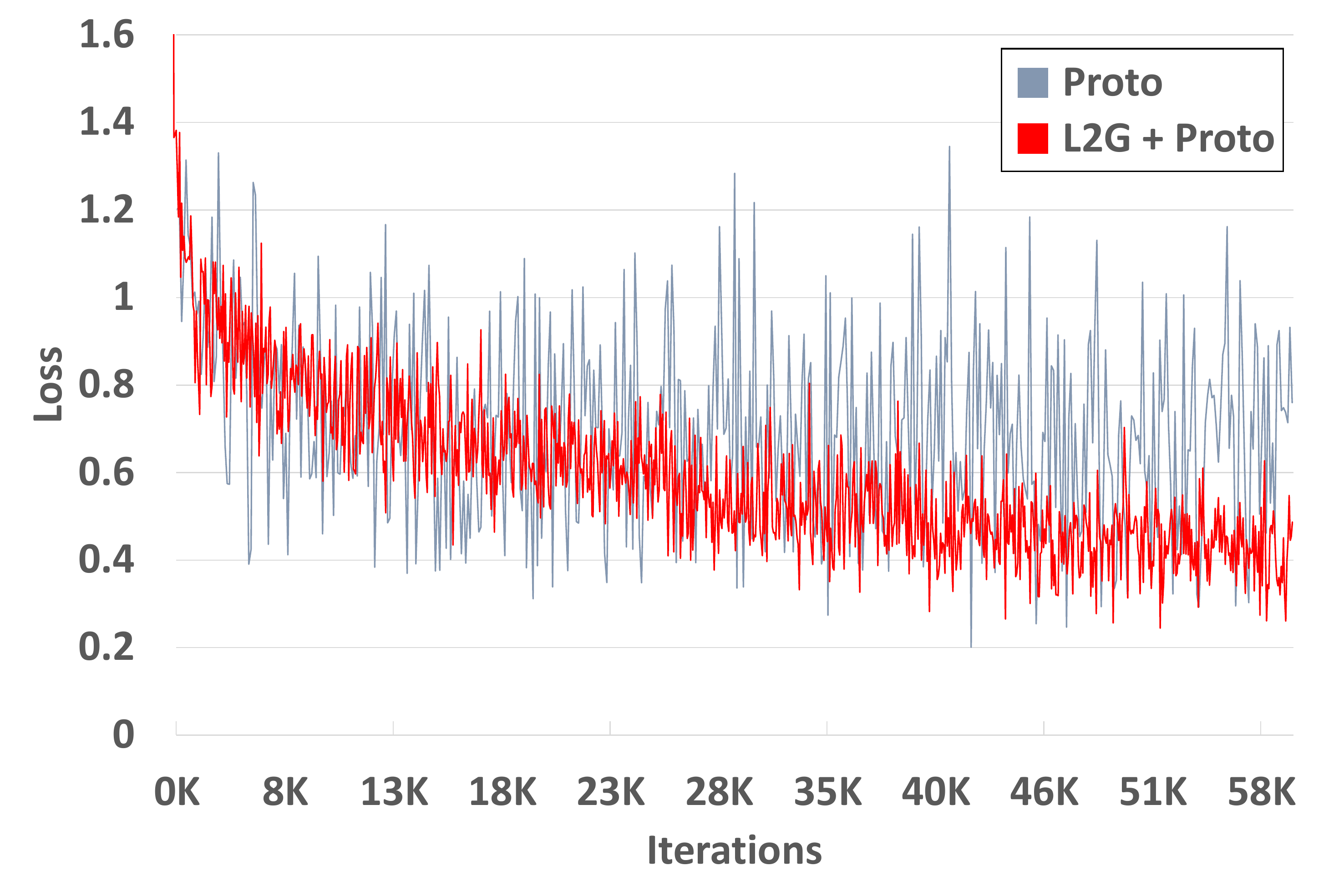}\label{fig:loss-mtr}}
	\subfigure[tiered-ImageNet]{\includegraphics[height=4cm]{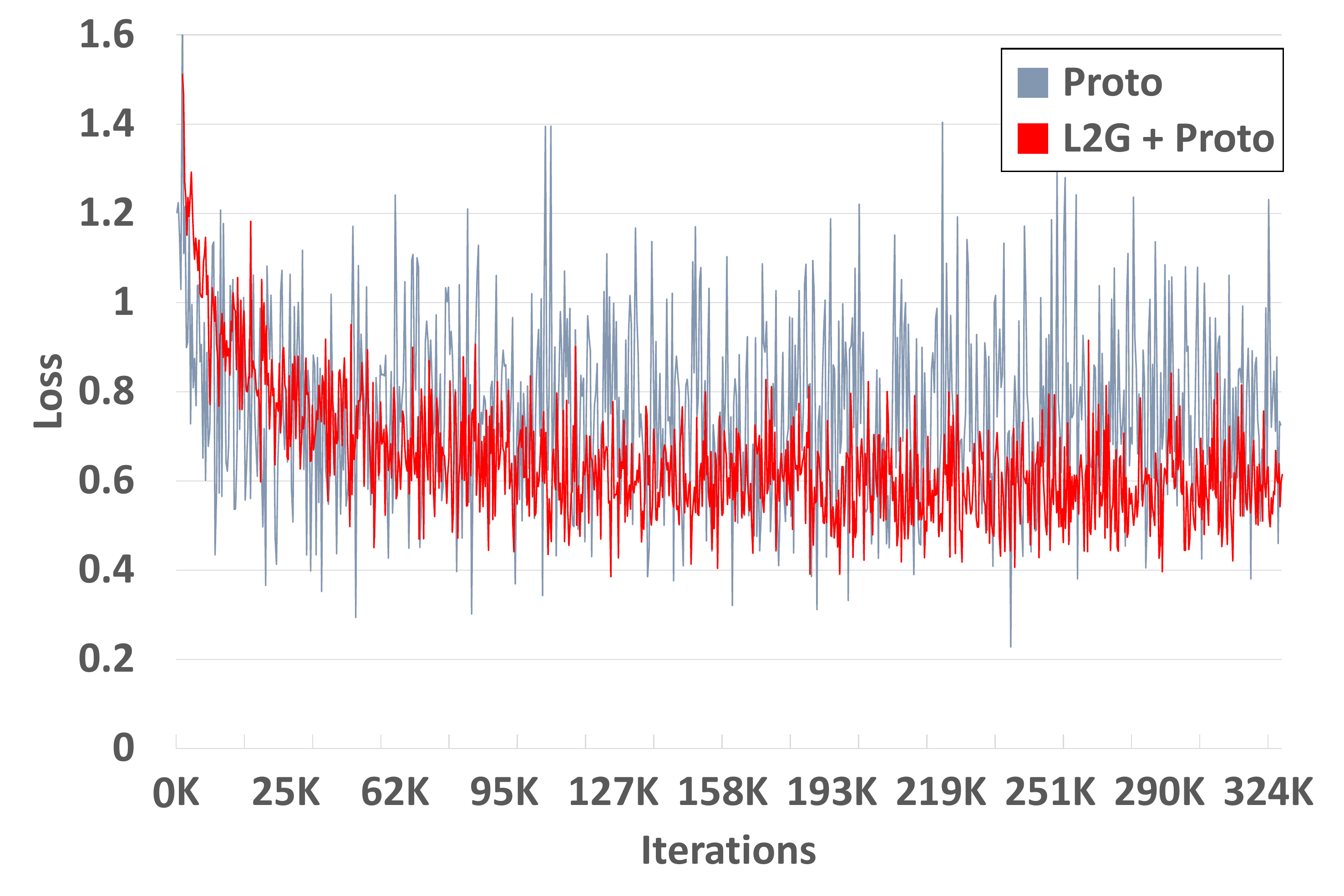}\label{fig:loss-ttr}}
	\vspace{-0.1in}
	\caption{\small \textbf{Convergence Plots.} Convergence plots on \textit{mini}-ImageNet and \textit{tiered}-ImageNet under 5-way 5-shot.}\label{fig:convergence}

\end{figure*}
\begin{figure*}[t]
	\centering
	\hfill
	\subfigure[Prototypical Networks.]{\includegraphics[height=3.3cm]{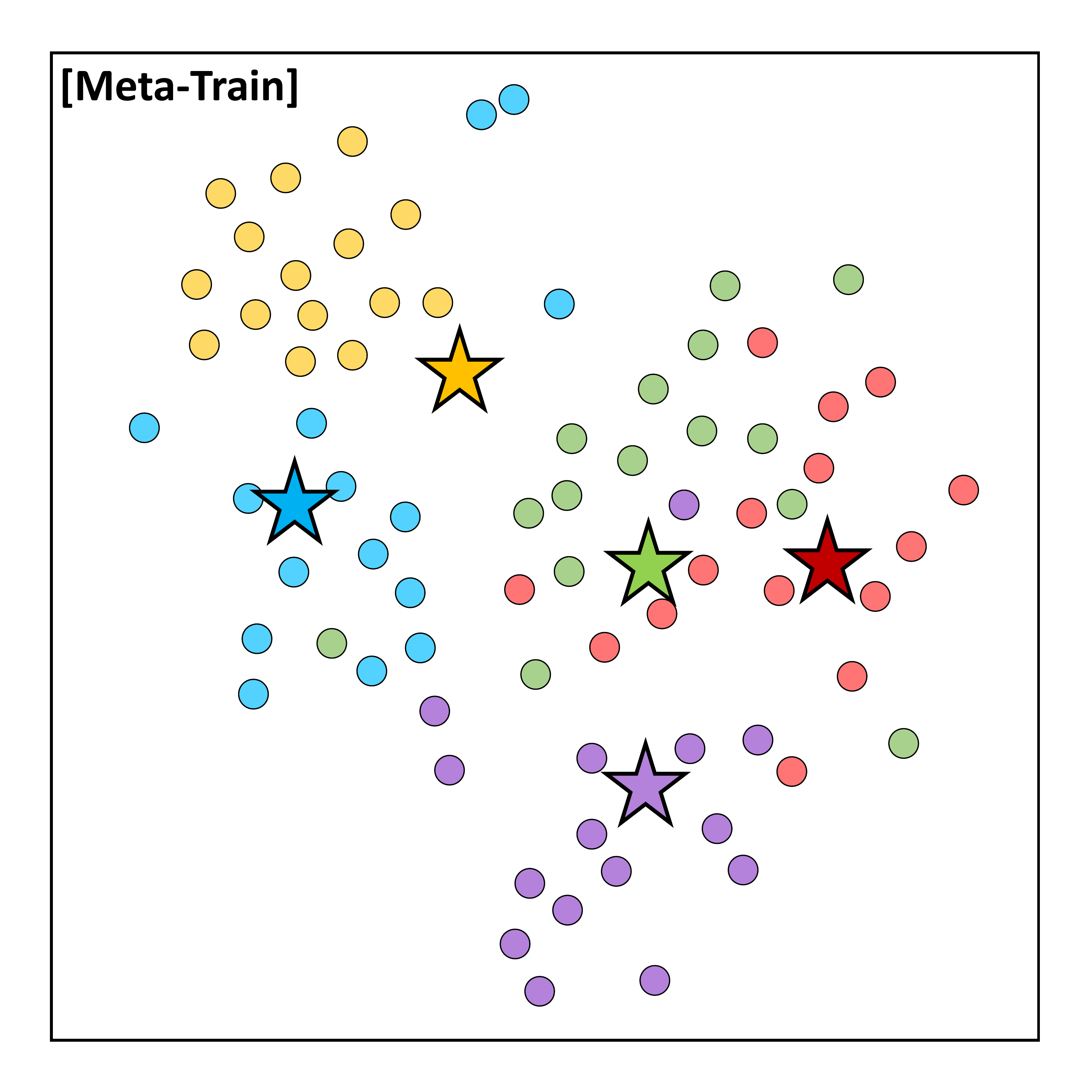}\label{fig:aproto_tr}
										\includegraphics[height=3.3cm]{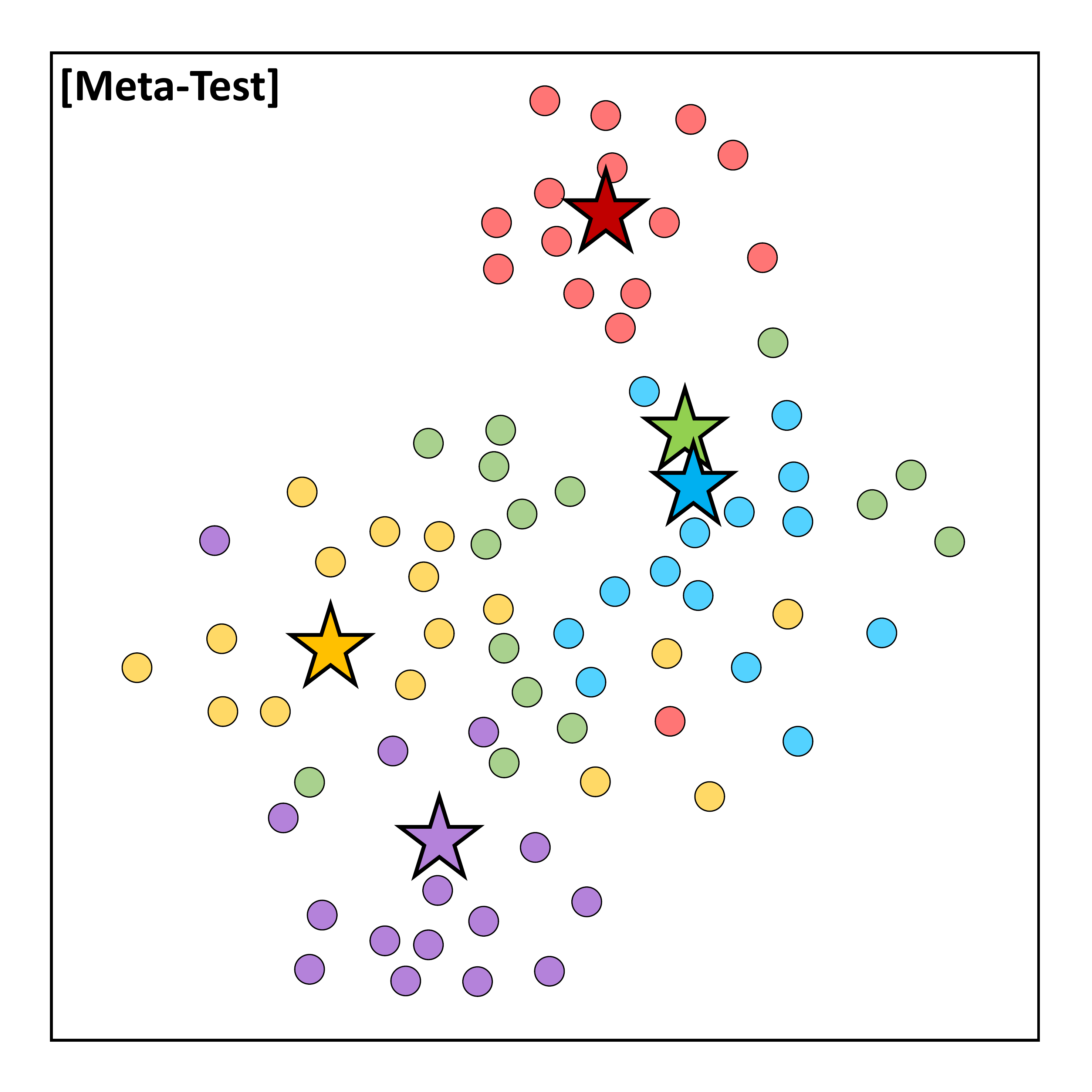}\label{fig:aproto_te}}
	\hfill
	\subfigure[L2G + Prototypical Networks.]{\includegraphics[height=3.3cm]{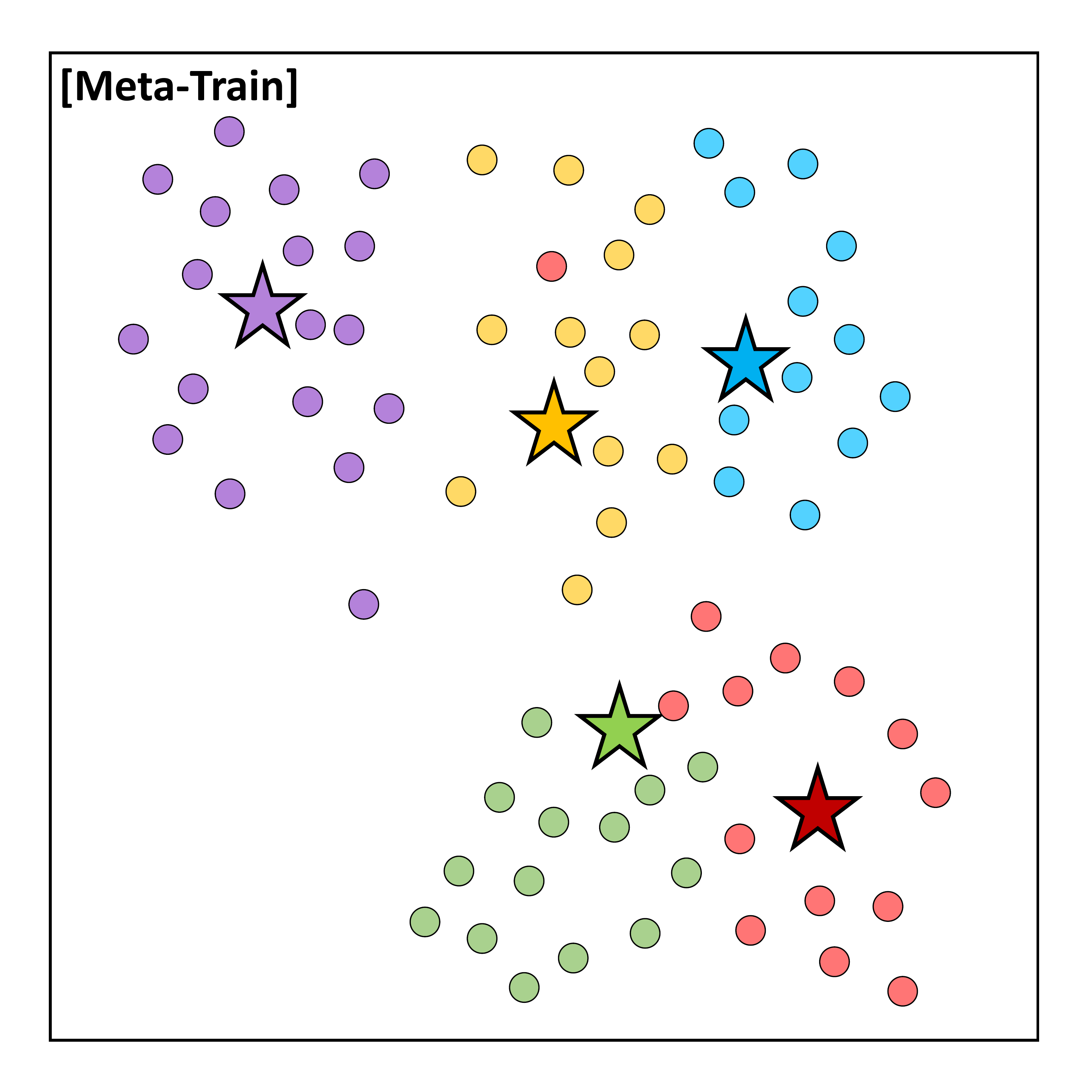}\label{fig:aours_tr}
											\includegraphics[height=3.3cm]{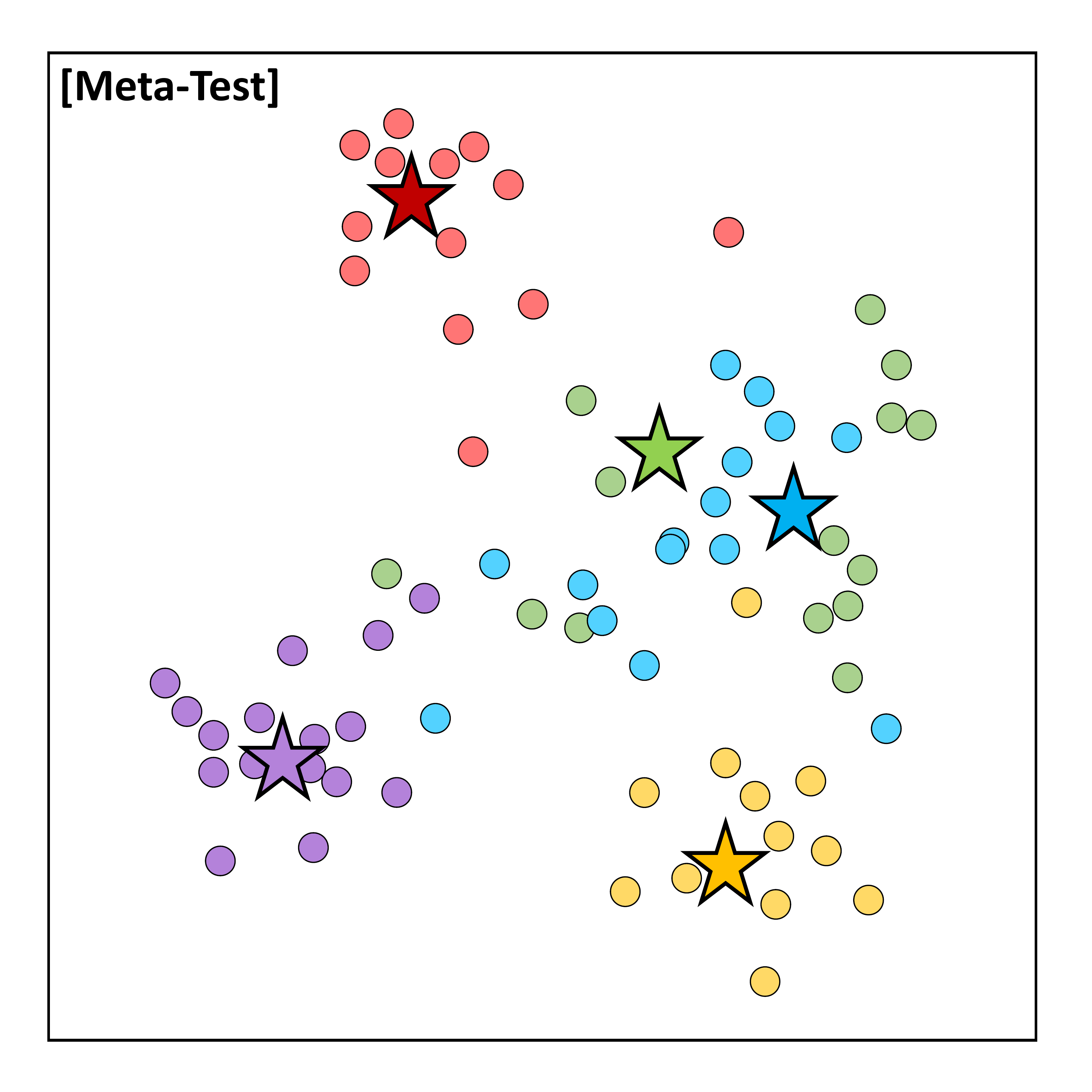}\label{fig:aours_te}}
	\hfill
	\vspace{-0.1in}
	\caption{\small \textbf{Embedding space visualization.} t-SNE visualizations of the learned embeddings on tiered-ImageNet. The stars denote support sets, and the colored dots denote query points for each of the classes. For both (a) and (b), the figure on the left is the visualization of the embedding after meta-training, and the figure on the right is the visualization of the embedding at the meta-testing time. We see that L2G obtains a space that separates well between unseen classes at meta-test time.}\label{fig:vis}
	\vspace{-0.1in}
\end{figure*}
We provide few-shot classification result under 5 and 10-way settings of \textit{mini}-ImageNet in Table~\ref{tbl:mini} and \textit{tiered}-ImageNet in Table~\ref{tbl:tiered} respectively. The results clearly show the effectiveness of L2G framework, since both models trained with our meta-learning framework outperform the baselines across all conditions. Specifically, when it comes to 1-shot classification, which has insufficient information for the inference, the L2G framework yields even larger performance improvements, with at most 4.89\% gain for 5-way 1-shot on \textit{tiered}-ImageNet over the baseline. These are impressive performance gains, considering the simplicity of the learning framework. MAML + ProtoTypical Networks and MAML + RelationNetworks models do now show noticeable improvements except for 10-way 5-shot tasks on \textit{mini}-ImageNet comparing to the original models which are Prototypical Network and Relation Net. This suggest that simple combinations between MAML and embedding models are not effective to learn generalization to unseen classes. 

For further analysis of the behavior of our L2G meta-learning, we show the convergence plot in Figure~\ref{fig:convergence}, which shows the meta-training loss. Models trained with our L2G framework converge only slightly slower compared to the base models, and eventually converge at lower loss. We also provide the visualizations of the learned embeddings in Figure~\ref{fig:vis}, which confirms the effectiveness of our framework. While the embeddings of the supports and queries from the with the original Prototypical Networks look scatters and overlapping, the model with L2G shows clean separation between classes. This may be due to its training objective whose main objective is to work well on unseen tasks.

\paragraph{\textit{Any}-way \& \textit{Any}-shot Classification}

\begin{table}[t]
	\small
	\caption{\small \textbf{\textit{mini}-ImageNet Any-way Any-shot classification.} The reported numbers are mean and standard errors with $95\%$ confidence interval over $5$ runs. Each run consists of the mean accuracy over $600$ episodes.} 
	
	\begin{center}
		\begin{tabular}{cccc}
			\hline
			Models & \textit{Any}-shot & \textit{Any}-way & \textit{Any}-shot \& \textit{Any}-way \\
			\hline
			\hline
			Prototypical Networks & 58.52 $\pm$ 0.38 & 57.94 $\pm$ 0.25 & 51.02 $\pm$ 0.30 \\
			\bf L2G + Prototypical Networks  & \bf60.64 $\pm$ 0.31 & \bf58.08 $\pm$ 0.48  & \bf53.00 $\pm$ 0.28 \\
			\hline
		\end{tabular}\label{tbl:mini-any}
		\vspace{-0.15in}
	\end{center}
\end{table}

To validate the leverage of generic information transfer, we introduce a novel \textit{any}-way \textit{any}-shot classification task, where both the number of shots and classes in each episode can largely vary. Table~\ref{tbl:mini-any} shows the results of the base model and ours on the \textit{mini}-ImageNet dataset. For any-shot classification, the number of shot randomly varies between ${1,5,10}$-shots for each episode, and simliarly, the number of classes at each episode randomly varies between ${5,7,10}$-ways for any-way setting. For training our L2G models, we fix the first task to a 5-shot classification problem and the generalization task to contain any-shot any-way classification problems. In the same manner, we train the Prototypical Network for any-shot and any-way classification. In the meta-test phase, we iteratively change the number of shots between ${1, 5, 10}$ and ways between ${5, 7, 10}$ with 600 episodes. Table~\ref{tbl:mini-any} shows the results. We observe that combining L2G framework with the model consistently improves over the base model, suggesting that it learned to generalize to unseen tasks with varying number of classes and instances. 

\section{Conclusion}
We proposed a novel meta-learning framework which we refer to as \emph{Learning to Generalize (L2G)}, which constrains the meta-learning process such that training on one task should generalize to unseen tasks. Specifically, we proposed a bilevel optimization problem, where we solve for a task by optimizing with respect to the task-shared parameter that should generalize well to another task. This framework goes well with metric-based meta-learning models, which learns a space that generalizes over any classification tasks. Based on this observation, we combine our model with two metric-based models, namely Prototypical Networks and Relation Networks. The models combined with our L2G framework significantly outperforms the base models when validated on standard few-shot classification tasks with fixed number of support instances, as well as on a novel task with varying number of support shots or classes. Further analysis with the embedding space visualization and the convergence plot shows the effectiveness of the \textit{learning to generalize} framework, as it provides clear separation between unseen classes when applied to a meta-test time and allows the model to converge to much lower test loss. 

\bibliography{neurips_format}
\bibliographystyle{neurips_2019_author_response}

\end{document}